\documentclass[10pt,twocolumn,letterpaper]{article}

\usepackage[pagenumbers]{cvpr} 
\usepackage{comment}
\usepackage{siunitx} 

\definecolor{cvprblue}{rgb}{0.21,0.49,0.74}
\usepackage[pagebackref,breaklinks,colorlinks,allcolors=cvprblue]{hyperref}

\def\paperID{14463} 
\def\confName{CVPR}
\def\confYear{2026}

\title{YOPO-Nav: Visual Navigation using 3DGS Graphs from One-Pass Videos}

\author{Ryan Meegan\\
Rutgers University\\
{\tt\small ryan.meegan@rutgers.edu}
\and
Adam D'Souza\\
Rutgers University\\
{\tt\small adam.dsouza@rutgers.edu}
\and
Bryan Bo Cao\\
Stony Brook University\\
{\tt\small boccao@cs.stonybrook.edu}
\and
Shubham Jain\\
Stony Brook University\\
{\tt\small jain@cs.stonybrook.edu}
\and
Kristin Dana\\
Rutgers University\\
{\tt\small kristin.dana@rutgers.edu}
}

\begin{document}
\maketitle
\begin{abstract}

Visual navigation has emerged as a practical alternative to traditional robotic navigation pipelines that rely on detailed mapping and path planning. However, constructing and maintaining 3D maps is often computationally expensive and memory-intensive. We address the problem of visual navigation when exploration videos of a large environment are available. The videos serve as a visual reference, allowing a robot to retrace the explored trajectories without relying on metric maps. Our proposed method, YOPO-Nav (You Only Pass Once), encodes an environment into a compact spatial representation composed of interconnected local 3D Gaussian Splatting (3DGS) models. During navigation, the framework aligns the robot’s current visual observation with this representation and predicts actions that guide it back toward the demonstrated trajectory. YOPO-Nav employs a hierarchical design: a visual place recognition (VPR) module provides coarse localization, while the local 3DGS models refine the goal and intermediate poses to generate control actions. To evaluate our approach, we introduce the YOPO-Campus dataset, comprising \ensuremath{\sim}4 hours of egocentric video and robot controller inputs from over \SI{6}{\kilo\meter} of human-teleoperated robot trajectories. We benchmark recent visual navigation methods on trajectories from YOPO-Campus using a Clearpath Jackal robot. Experimental results show YOPO-Nav provides excellent performance in image-goal navigation for real-world scenes on a physical robot. The dataset and code will be made publicly available for visual navigation and scene representation research. 

\end{abstract}    
\vspace{-0.65cm}
\section{Introduction}
\label{sec:intro}

\begin{figure}[t!]
\centering
\includegraphics[width=0.48\textwidth, height=0.19\textheight]{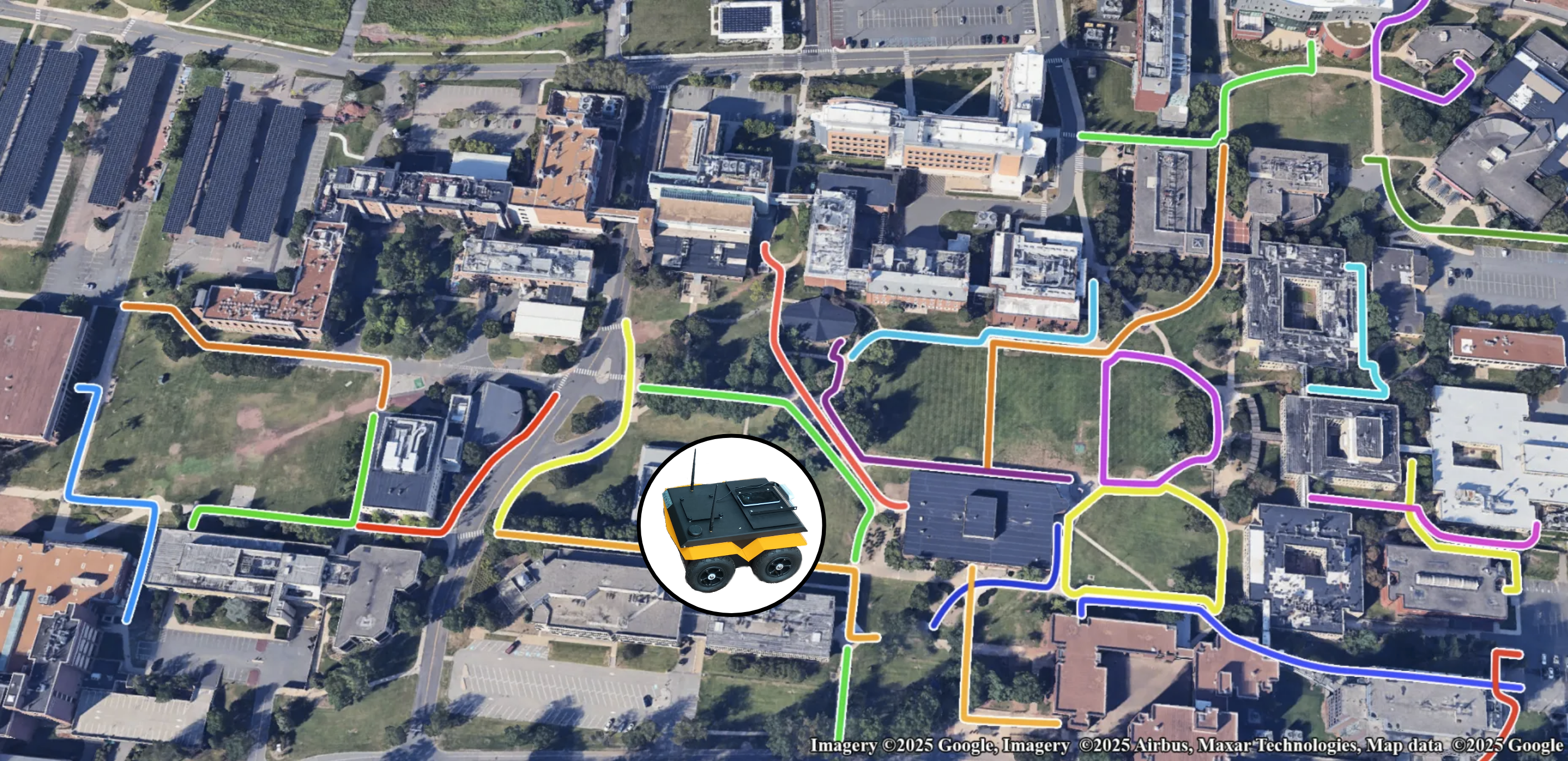}
\caption{
{\bf YOPO-Campus Dataset} Birds-eye-view (BEV) of the paths traversed by the Jackal robot under human teleoperation across Rutgers University, Busch Campus. Egocentric video from the robot and action control sequences are captured over 4 hours across \SI{6}{\kilo\meter}.
}
\label{fig:yopo_campus_bev}
\end{figure}

In unfamiliar environments, humans naturally explore to learn the scene, forming a mental map organized around key landmarks, salient objects, topology, and spatial relationships \cite{chrastil2014cognitive, peer2021structuring, weisberg2016some}. The concept of cognitive mental maps has driven new computational scene representations and methods for {\it visual navigation}, enabling navigation on low-cost mobile robots equipped with only an onboard camera and no odometry, GPS, or multi‑modal sensing.

Traditionally, robotic navigation is treated as a geometric problem: exhaustively traversing the environment to build a detailed 3D reconstruction, then planning using the generated representation. This large-scale mapping approach is computationally intensive and often impractical for time-sensitive applications, where prolonged exploration is not feasible. Moreover, human activity in crowded settings further complicates robot exploration and mapping. Unlike humans, who enter new environments with prior knowledge of scene semantics and appropriate movement patterns, robots using traditional SLAM lack this higher-level contextual awareness.

\begin{figure*}[t!]
\centering
\includegraphics[width=0.93\textwidth, height=0.42\textheight]{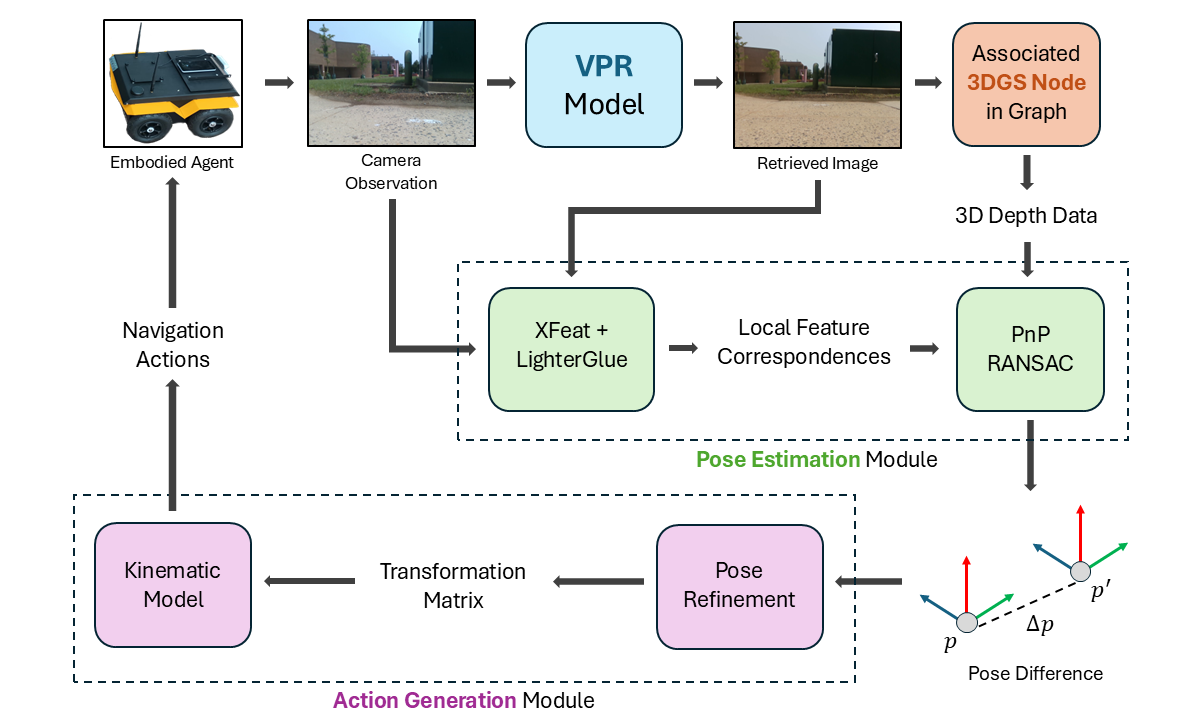}
\caption{
\textbf{YOPO-Nav:} Videos of human-teleoperated robot trajectories are used to construct a scene representation as a graph of local 3DGS nodes. When the embodied agent revisits the scene, coarse localization is performed using Visual Place Recognition (VPR), linking frames in the the recorded trajectory to a corresponding node in the 3DGS graph. The robot’s real-world pose, $p'$, is localized in the 3DGS node via PnP RANSAC, and the difference from the desired pose, $p$, yields a transformation matrix, directing actions to align with $p$ (see Section~\ref{subsec:yoponav}.)
}
\label{fig:yoponav}
\end{figure*}

Early efforts to leverage visual information for scene representations primarily employed topological graphs \cite{gupta2017cognitive, chang2020semantic, shah2021ving, shah2022gnm, shah2023vint, sridhar2024nomad}. Recent approaches leverage generative models to anticipate future trajectories, enabling more accurate navigation decisions \cite{bar2025navigation, zhang2024imagine, koh2021pathdreamer}. 
Currently, 3D Gaussian Splatting (3DGS) models \cite{kerbl20233d} are changing visual navigation by supporting real-time, photo-realistic rendering, and novel view synthesis for algorithmic navigation decisions. Scene representations with large global 3DGS models are a shift from both classic SLAM and topological graphs \cite{tosi2024nerfs, matsuki2024gaussian, yan2024gs}.

Building 3D scene representations directly from one-pass videos is highly compelling, as videos are often easy to capture \cite{chang2020semantic, xie2025vid2sim, schmeckpeper2020reinforcement, liu2024citywalker, bar2025navigation}. However, a key limitation of videos is that they lack true 3D structure: the scene is entirely constrained to the trajectory originally taken. Since 3D Gaussian Splatting can transform a sequence of video frames into Gaussian primitives that encode depth and interpolate radiance across viewpoints, it allows free movement along the recorded trajectory and exploration beyond the original perspectives. In other words, a robot or agent that strays from the recorded trajectory can leverage the constructed 3D representation to compute geometry‑based corrective actions and realign with the trajectory. This insight motivates our approach: an exploration video defines a traversable path through the scene, a “video breadcrumb trail”, that the agent can later follow to navigate the environment.

\begin{figure}
\centering
\includegraphics[width=0.44\textwidth, height=0.20\textheight]{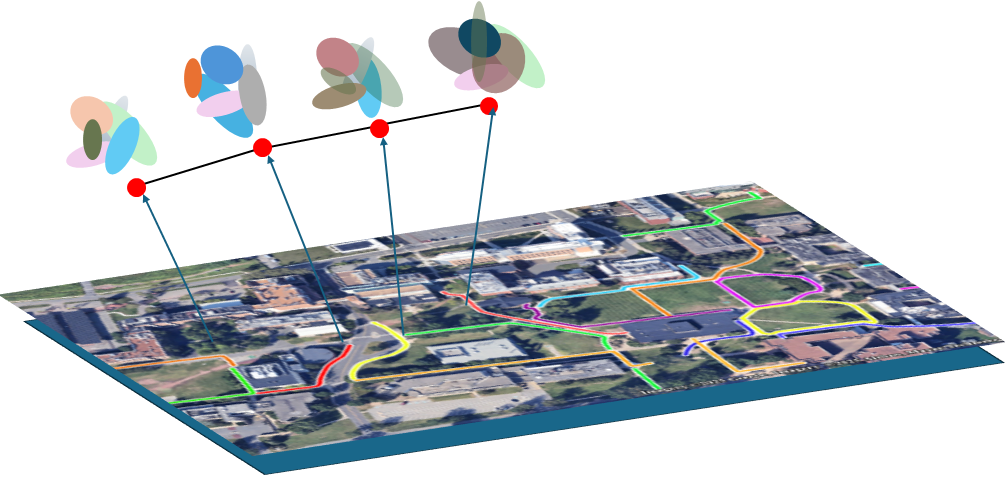}
\caption{
\textbf{YOPO-Nav Scene Representation} YOPO-Nav represents a scene as a graph of 3DGS models, built from \ensuremath{\sim}50–55 frames at $448 \times $336 resolution, using videos of human-teleoperated robot trajectories. Edges connect nodes by frame continuity (within each video) or by visual similarity (across different videos). Navigation proceeds by localizing new camera observations in the 3DGS and aligning them to the estimated poses from the videos.}
\label{fig:yoponav_rep}
\end{figure}

In this work, we merge the concepts of topological graphs and 3DGS methods to construct a graph of local 3DGS models from one-pass videos. We propose a framework called YOPO-Nav (You Only Pass Once), encoding an environment into a compact spatial representation composed of interconnected local 3D Gaussian Splatting (3DGS) models (see Figs.~\ref{fig:yoponav} and \ref{fig:yoponav_rep}). Our approach is lightweight, interpretable, and scalable, relying only on single-pass videos of the scene; no GPS or odometry is assumed. The YOPO-Nav scene representation draws inspiration from human mental maps: identifying points of interest, partitioning the environment into manageable chunks, and organizing these chunks into an interconnected structure (see Fig.~\ref{fig:yoponav_rep}). Navigation proceeds hierarchically: Visual Place Recognition (VPR) localizes the start and goal regions, recalls the surroundings, and links connecting chunks into a traversable, continuous route.

Key to our approach is the integration of frameworks that can create 3DGS models from un-calibrated multiview images \cite{jiang2025anysplat, zhang2025flare, kang2025selfsplat, Hong_2025_PF3plat, lin2025longsplat}. By jointly estimating 3D Gaussians and poses, we obtain a stable representation that enables scene learning from one-pass videos and supports navigation on a physical robot (Clearpath Jackal). 

As a contribution of this paper, and to test YOPO-Nav, we introduce YOPO-Campus (see Fig.~\ref{fig:yopo_campus_bev}), a dataset collected on a college campus using a Clearpath Jackal robot equipped with an Intel RealSense D435 camera and teleoperated by a human via a wireless DualShock controller. The robot traversed 35 unique trajectories along sidewalks connecting campus buildings, covering \SI{6}{\kilo\meter} (4 hours of footage). We evaluate YOPO-Nav in the real world on YOPO-Campus trajectories, demonstrating excellent performance in visual navigation on the common task of image-goal navigation, without reliance on odometry and GPS. We further demonstrate strong transfer of pre-trained VPR networks, where a geographically distinct and unaffiliated dataset (GND: Global Navigation Dataset \cite{liang2025gnd}) is used for pre-training but tested on the YOPO-Campus trajectories. 

Our contributions are: (i) YOPO-Campus, a \SI{6}{\kilo\meter} (4 hours) dataset collected by a human-teleoperated robot in an outdoor college campus environment  (ii) Scene representation comprised of a \textit{graph of 3D Gaussian Splats}, where small spatial regions are represented by their own local 3DGS models (iii) YOPO-Nav, an algorithm utilizing this graph of 3DGS models that enables visual navigation across scenes with minimal human intervention.

\vspace{-0.15cm}
\section{Related Work}
\label{sec:relatedworks}

Our work relates to four main lines of research: (1) Scene Representations, (2) 3D Gaussian Splatting, (3) Visual Place Recognition, and (4) Embodied Visual Navigation.

\vspace{-0.38cm}
\paragraph{\bf Scene Representations for Navigation} Beyond full metric maps, alternative scene representations have been introduced to learn navigation policies. Scene graphs represent visual environments as structured graphs, with nodes denoting objects and edges capturing semantic or spatial relationships. Scene graphs, like topological graphs, are typically non‑metric; but while topological graphs map spatial regions and connectivity, scene graphs span broader entities such as objects, people, and regions. Recent work \cite{werby2024hierarchical, singh2023scene} extends scene graphs for navigation by encoding both object relations and spatial connectivity.

Bird’s-eye-view (BEV) representations provide an alternative using occupancy grids or similar structures to capture metric spatial relations. BEV scene graphs \cite{liu2023bird} unify these ideas by constructing a scene graph on top of a BEV representation, combining BEV’s geometric grounding with the relational reasoning of scene graphs. Topological graphs have evolved from early cognitive-map-inspired models, which were symbolic abstractions of connected places, to recent approaches \cite{singh2023scene, loo2025open, johnson2024feudal, kim2023topological, li2024memonav} that integrate learned spatial embeddings, semantic grounding, and semantic augmentation. Our method models the scene as a graph that captures spatial relationships without encoding global metrics, assigning each node to a local 3D Gaussian Splat.

\vspace{-0.35cm}
\paragraph{\bf 3D Gaussian Splatting} 3D Gaussian splatting \cite{kerbl20233d} has revolutionized scene modeling. Unlike traditional SLAM, which outputs point clouds that are later converted to surfaces, 3DGS captures the true visual appearance of a scene using efficient geometric models composed of Gaussian primitives. 3DGS methods rasterize Gaussian primitives iteratively, producing high‑quality 3D models with novel view synthesis. By leveraging established rasterization techniques, 3DGS methods achieve substantially faster rendering than NeRF \cite{mildenhall2021nerf}. With recent advancements in 3D Gaussian Splatting, the technology has been applied to multiple downstream tasks, including navigation. 3DGS has been integrated into SLAM algorithms to construct detailed volumetric maps of the environment for localization and planning \cite{matsuki2024gaussian, yan2024gs, li2024sgs, tosi2024nerfs}. Recent work applies Gaussian Splatting to embodied visual navigation for fast novel view synthesis \cite{lei2025gaussnav} and pose estimation \cite{chen2025splat, guo2025igl}. Building on recent advances in 3DGS, we avoid global modeling and instead propose a graph framework of local 3DGS nodes, combining the strengths of topological graphs and 3DGS models.

\vspace{-0.35cm}
\paragraph{\bf Visual Place Recognition} The goal of visual place recognition (VPR) \cite{arandjelovic2016netvlad} is to determine whether an image has been seen before, and this framework is useful in robot localization and navigation. Modern VPR models typically combine a deep learning feature extractor with an aggregator. Common optimizations include enhancing the feature extractor \cite{ali2023mixvpr, keetha2023anyloc}, reformulating aggregator clustering \cite{izquierdo2024optimal}, and improving viewpoint robustness \cite{berton2023eigenplaces}. These advances have enabled the use of VPR in various embodied robotic navigation tasks \cite{claxton2024improving, garg2021your}. VPR not only provides an effective method for localization, but can also serve as a foundation for representing the environment in navigation tasks. We build on recent VPR advancements to identify the node in the 3DGS graph where the agent is currently located. Our VPR model achieves strong performance when trained on a domain-similar (but geographically distinct) dataset, GND: Global Navigation Dataset \cite{liang2025gnd}.

\vspace{-0.38cm}
\paragraph{\bf Embodied Visual Navigation} Several tasks have been proposed in embodied visual navigation, including object-goal navigation \cite{sun2024survey}, multi-object-goal navigation \cite{wani2020multion, chen2022learning}, image-goal navigation \cite{zhu2017target, chaplot2020neural, kwon2021visual, kwon2023renderable}, and instance-image-goal navigation \cite{krantz2022instance, bono2023end}, with most evaluated in simulation environments such as Habitat \cite{savva2019habitat} and Gibson \cite{xia2018gibson}. Deploying visual navigation algorithms on physical robots remains a core challenge, and key objectives are high success rates for tasks in real environments.

Various strategies have been applied to embodied visual navigation, including deep reinforcement learning \cite{zhu2017target, lobos2018visual, devo2020towards}, spatial attention \cite{mayo2021visual, seymour2021maast, lyu2022improving}, and novel view synthesis \cite{koh2021pathdreamer}. Recent approaches also employ diffusion models to sample from the action space \cite{ren2025prior} or to predict future observations along a trajectory \cite{bar2025navigation, zhang2024imagine} to determine optimal navigation actions. ViNT \cite{shah2023vint} introduced a foundation transformer model for visual navigation that uses diffusion to predict intermediate goal images and a topological graph structure for mapping and planning. Several subsequent works \cite{shah2022gnm, sridhar2024nomad} proposed similar generalizable navigation policies, combining transformer architectures with diffusion models. Unlike generalist navigation foundation models, our method focuses on learning a specific environment from a single exploration video. Our results indicate that YOPO-Nav has better performance than pre-trained foundation models, suggesting that when one or more exploration videos are available, a strong representation can be produced.
\vspace{-0.18cm}
\section{Methods}

\subsection{YOPO-Campus}
\label{subsec:yopo-campus}
YOPO‑Campus, our dataset of robot trajectories, was collected at Rutgers University, Busch Campus, during the Summer and Fall of 2024 and 2025, primarily in low‑activity periods (summer break, weekends, or early mornings) to ensure that no people appear in the videos. The dataset consists of timestamps, synchronized RGB (8‑bit) and depth (16‑bit) imagery at VGA resolution ($640 \times 480$), controller inputs, compass directions, and ground‑truth GPS and Wi‑Fi Fine Timing Measurements (FTM). Data was gathered using a Clearpath Jackal robot, equipped with an Intel RealSense D435 camera and a Google Pixel 3a smartphone, that was manually teleoperated by a human using a wireless DualShock controller. The Pixel 3a, mounted on the robot and connected via Bluetooth to the Jackal robot, logged GPS, compass direction, and Wi‑Fi measurements (FTM and RSSI) through a custom Android app we developed. To support the collection of FTM and RSSI data, 3 Google Nest Pro routers were placed along each trajectory, and their GPS coordinates were recorded. Wi-Fi FTM and RSSI measurements are intended for future research and are currently not used in YOPO-Nav. GPS is used only for evaluation of our VPR model.

Controller inputs were limited to discrete actions: \SI{0.25}{m} forward and backward using the up and down arrows on the D-pad, and \(\pm 15^{\circ}\) rotations using the square and circle buttons, respectively. After every action, a timestamp is logged, RGB and depth images are captured, and the Pixel 3a is queried for GPS, compass direction, and Wi‑Fi FTM and RSSI data, which is sent back to the Jackal robot over a Bluetooth connection. Data was collected along 35 sidewalk trajectories (\SI{100}{m}–\SI{250}{m}, avg. \SI{170}{m}), each with unique start and end points. Trajectories were traversed bidirectionally to capture complementary perspectives, lasting 3–16 minutes (avg. 6.5 minutes) per path. The dataset covers most of the campus sidewalk network and includes academic buildings, statues, and open fields. At sidewalk junctions, the Jackal robot was rotated in place by the human teleoperator to more easily align overlapping path segments. In total, the dataset contains 4 hours of data (26,500 images, actions, etc.) that covers \SI{6}{\kilo\meter} of the campus. The dataset has a file size of 26.5 GB (13.1 GB with compression). A Qt-based GUI was created to view the dataset in an efficient manner (see Fig.~\ref{fig:yopo_campus_dataset_viewer}).

\begin{figure}[t!]
\centering
\includegraphics[width=0.48\textwidth, height=0.15\textheight]{sec/figures/dataset_viewer.png}
\caption{
\textbf{YOPO-Campus Dataset Viewer} GUI for efficient visualization of YOPO‑Campus: left: bird’s‑eye view annotated with routers, planned path, and robot position based on the current frame; right: RGB/depth images; center: frame data (timestamp, action, FTM/RSSI, compass, GPS) with a player to view the data associated with each frame in the video.}
\label{fig:yopo_campus_dataset_viewer}
\end{figure}

\vspace{-0.10cm}
\subsection{YOPO-Nav}
\label{subsec:yoponav}
YOPO-Nav integrates two complementary components. A visual place recognition model offers a coarse global estimate of location, while a graph of compressed 3DGS models (see Fig.~\ref{fig:yoponav_rep}) captures fine local geometry and appearance. Together they establish a global‑to‑local hierarchy for navigation (see Fig. \ref{fig:yoponav}): global image retrieval selects the relevant 3DGS, and local pose estimation determines the robot’s position within that 3DGS and computes the action needed to align with the estimated poses from the recorded trajectories. This representation is lightweight, avoiding the cost of full 3D reconstruction in SLAM approaches and loss of detail in topological graph approaches, while remaining interpretable for human guidance and correction.

\vspace{-0.32cm}
\paragraph{\bf VPR} The visual place recognition model, hereafter referred to as YOPO-Loc, was trained following the open-source framework created by OpenVPRLab \cite{OpenVPRLab}. This framework consists of a datamodule, feature backbone, aggregator, and loss module. DINOv3 \cite{simeoni2025dinov3}, the current state-of-the-art feature extractor, was chosen as the backbone while Bag of Queries (BoQ) \cite{ali2024boq} was chosen as the aggregator due to its strong performance and deployment simplicity \cite{ali2024boq, wang2025focus}. Multi-similarity loss was selected as the loss module because of its stable training behavior. The Global Navigation Dataset (GND) \cite{liang2025gnd} was selected as the datamodule. GND is a large-scale dataset collected across ten university campuses, covering approximately \SI{2.7}{\square\kilo\metre} and containing RGB images and associated GPS. We selected GND for training because it resembles our dataset (a Jackal robot navigating a campus environment) and we expected successful transfer learning. 

\begin{table}
  \centering
  \scriptsize
  \setlength{\tabcolsep}{3pt}
  \scalebox{1.2}{
    \begin{tabular}{@{}lccccc@{}}
      \toprule
      Dataset & R@1 ↑ & R@5 ↑ & R@10 ↑ & R@15 ↑ & $\Delta$R@1 ↑ \\
      \midrule
      GND Val Set \cite{liang2025gnd}    & 92.27 & 97.16 & 98.37 & 98.64 & +7\% \\
      YOPO-Campus & 68.54 & 93.68 & 96.55 & 97.37 & +3\% \\
      \bottomrule
    \end{tabular}
  }
  \caption{Recall results of YOPO-Loc on YOPO-Campus and the GND \cite{liang2025gnd} validation set. R@1 improved by 3\% on YOPO‑Campus and 7\% on the GND validation set. The performance gain on YOPO‑Campus provides evidence of transfer learning from GND. The R@1–R@5 gain on YOPO‑Campus motivates a top‑5 KNN search in the FAISS \cite{douze2025faiss} index built from YOPO‑Loc features.}
  \label{tab:vpr_results}
\end{table}

\begin{table}
  \centering
  \scriptsize
  \setlength{\tabcolsep}{2pt}
  \scalebox{1.02}{
    \begin{tabular}{@{}lccccccc@{}}
      \toprule
      Model & Params (M) ↓ & FLOPs (G) ↓ & R@1 ↑ & R@5 ↑ & R@10 ↑ & R@15 ↑ \\
      \midrule
      SALAD \cite{izquierdo2024optimal} & \textbf{87.99} & 22.22 & 67.89 & 92.95 & 96.26 & \textbf{97.41} \\
      MegaLOC \cite{berton2025megaloc} & 228.64 & 22.37 & 68.54 & 93.51 & 96.53 & 97.31 \\
      FoL \cite{wang2025focus} & 308.83 & 164.75 & 66.54 & 92.52 & 95.94 & 97.17 \\
      YOPO-Loc & 99.85 & \textbf{15.74} & \textbf{68.62} & \textbf{93.95} & \textbf{96.55} & 97.39 \\
      \bottomrule
    \end{tabular}
  }
  \caption{Comparison of visual place recognition models on YOPO-Campus. Best values in each column are in bold. YOPO-Loc achieves competitive recall compared to general-purpose models while requiring fewer parameters (with the exception of SALAD \cite{izquierdo2024optimal}) and FLOPs.}
  \label{tab:vpr_comparisons}
\end{table}

\begin{figure*}[t!]
\centering
\includegraphics[width=0.90\textwidth, height=0.3\textheight]{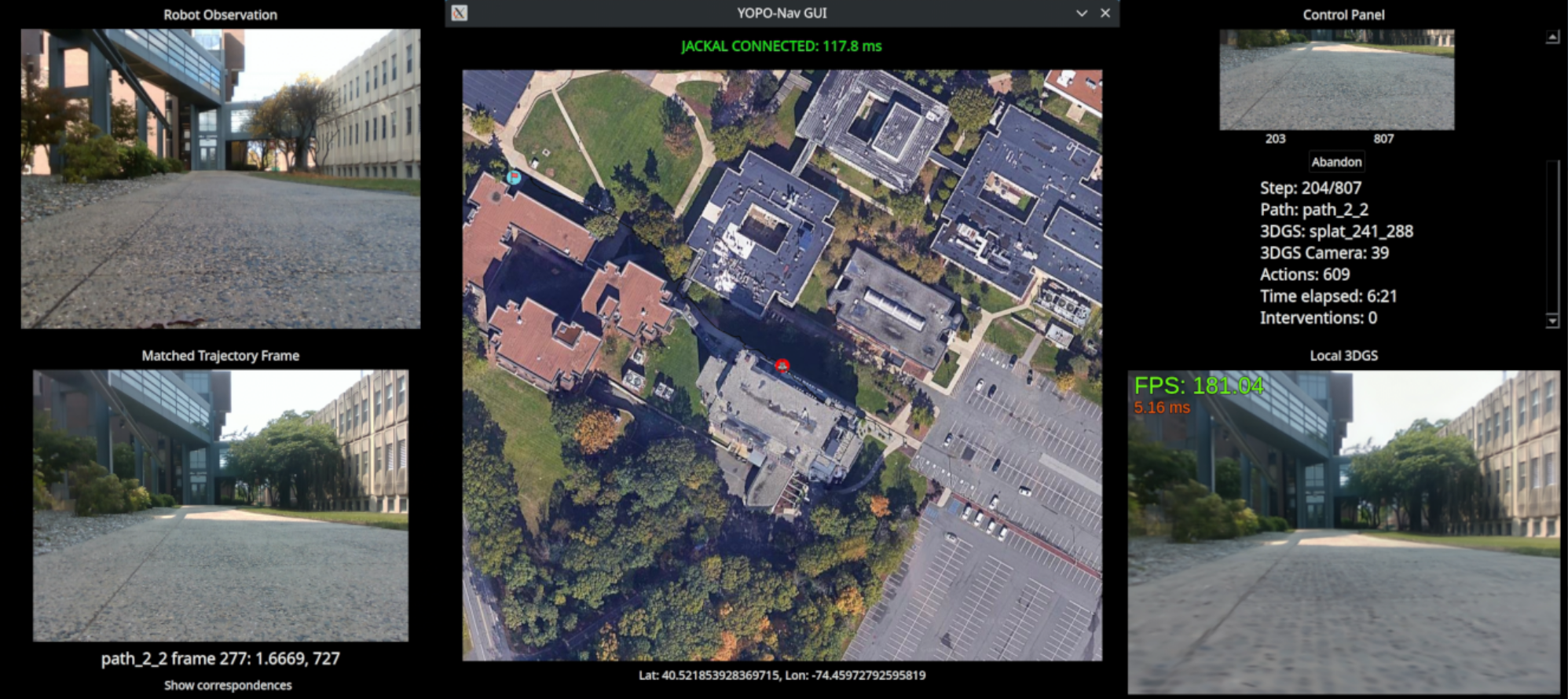}
\caption{
\textbf{YOPO-Nav GUI} The YOPO-Nav GUI is comprised of five widgets: 1) the top-left displays the Jackal robot's live camera feed; (2) the bottom-left shows the camera feed's closest matched frame in the FAISS \cite{douze2025faiss} index; (3) the center presents a BEV of the campus with the robot’s position, desired goal, and planned path; (4) the top-right displays the start and goal images, next frame in the planned path, and performance metrics (actions, time, interventions); and (5) the bottom-right renders the 3DGS model and simulated actions in real-time.}
\label{fig:yopo_nav_gui}
\end{figure*}

Only 6 of the 10 campuses in GND \cite{liang2025gnd} were used, since the others lacked GPS. After preprocessing, we split at the campus level, assigning all images from a given campus exclusively to either training (80\%) or validation (20\%) to prevent cross-contamination between sets. In total, there were 55,317 training and 14,584 validation images. Within each campus, images were defined as belonging in the same place if their associated GPS coordinates were within a \SI{1.0}{m} radius of each other. For validation, database images were defined as images within a \SI{0.5}{m} radius of each other, while query images were sampled from a \SI{0.5}{m} band outside each database region. Queries were matched against the database and performance was evaluated using recall@k (e.g., R@1, R@5), the percentage of queries that successfully retrieved the matching database within the top‑$k$ results. See Section~\ref{subsec:sota_comparisons} for training parameters.

YOPO‑Campus served as an additional validation set, following the same protocol as GND \cite{liang2025gnd}. Validation results for both sets are shown in Table~\ref{tab:vpr_results}. Evidence of transfer learning from GND to YOPO-Campus was demonstrated in YOPO-Loc. YOPO-Loc was compared to several notable VPR models on the YOPO-Campus dataset in Table~\ref{tab:vpr_comparisons}. YOPO-Loc achieves similar results to general-purpose models while being considerably more efficient. 

\vspace{-0.23cm}
\paragraph{\bf 3DGS} We employ AnySplat \cite{jiang2025anysplat}, an efficient feed‑forward network that generates 3D Gaussian primitives along with camera intrinsics and extrinsics in a single pass, to construct the graph of 3DGS models. We selected AnySplat because the lack of camera poses in our dataset requires a 3DGS method capable of estimating robust camera poses. We found that AnySplat produced the least sparse reconstruction with the most accurate camera pose estimations compared to other pose-free alternatives, such as VGGT \cite{wang2025vggt}, MapAnything \cite{keetha2025mapanything}, and LongSplat \cite{lin2025longsplat}. Frames from each trajectory in YOPO‑Campus were processed in AnySplat \cite{jiang2025anysplat} and exported as .ply files along with JSON files containing camera extrinsics, intrinsics, and the filenames of the frames used in the 3D reconstruction. See Section \ref{subsec:sota_comparisons} for more details. We converted all .ply files into the Spatially Ordered Gaussians (SOG) format using the PlayCanvas splat-transform library \cite{SplatTransform}, an enhanced implementation of Self-Organizing Gaussians \cite{morgenstern2024compact}, to reduce the storage footprint of all the model reconstructions. Across all 35 trajectories, 547 3DGS models were produced, with a total file size of 17.8 GB after SOG compression (JSONs total 30 MB). During runtime, SOG files are reconverted to .ply files, which takes 1-3 seconds, so they can be properly decoded and rendered by AnySplat.

All images in YOPO-Campus, contained within the 547 3DGS models, are used to built a graph that involves two stages: within each trajectory video, frames are connected sequentially based on the known order, and across different trajectory videos, visually similar frames are connected. For cross‑trajectory linking, YOPO‑Loc features for all frames in YOPO-Campus were stored in a GpuIndexFlat FAISS \cite{douze2025faiss} database, allowing each frame from one trajectory to retrieve its top‑5 candidates from other trajectories via a simple KNN search. The resulting top-5 candidates are re‑ranked using point correspondences from XFeat \cite{potje2024xfeat} and LighterGlue, an improved version of LightGlue \cite{lindenberger2023lightglue} by the authors of XFeat. An edge is retained if the image with the highest number of correspondences exceeds a predefined threshold (i.e. 500-900, 800 was determined to be optimal). The resulting graph supports path planning, e.g.\ with Dijkstra’s algorithm.

\vspace{-0.3cm}
\paragraph{\bf Pose Estimation and Action Generation} Navigation begins by loading the 3DGS model and pose for each frame in the planned path. Using the current camera observation, the Jackal robot is then localized within the 3DGS model. Translation and rotation errors between the estimated pose and the target pose (next frame in the planned path) are computed and corrective actions are executed to align with the target pose in 3DGS environment. These actions are transferred from the 3DGS environment into the real-world and are repeated for each frame in the planned path until the goal is reached. Two strategies enable this transfer: (1) deriving a scale factor from the camera height in the 3DGS node relative to its real‑world height to map translations in the 3DGS model to metric units, and (2) applying Perspective‑n‑Point (PnP) \cite{hartley2003multiple} RANSAC \cite{fischler1981random} to localize new camera observations within the 3DGS model.

Upon reaching a frame linked to a new 3DGS model, an image is rendered at each estimated camera pose (image inputs used in reconstruction) in the model. Segment Anything 2 \cite{ravi2024sam2segmentimages} (specifically multi-mask version of SAM‑2.1‑base‑plus) isolates the ground plane in each rendered image using a single point click positioned at 90\% of the image height and centered along the width. It is assumed that, across each image in YOPO-Campus, the ground planes will exist in this region—and it should exist based on our data collection procedure. The largest mask is always used to ensure the as much of the ground plane is segmented as possible. We define the ground plane as sidewalk, road, or brick pavement that forms the trajectory the Jackal robot follows. For each rendered image, the depth values within the ground‑plane mask are back‑projected into 3D points using the camera intrinsics, and a plane is fit to these points via RANSAC. The offset of each fitted plane is then compared to the Jackal robot’s fixed camera height of \SI{250}{mm} to compute a scale factor that converts 3DGS translations into real‑world meters (rotations require no scaling). The final scale factor is obtained by taking the median across the plane offsets for each estimated camera pose in the 3DGS model.

To localize within the 3DGS model, feature correspondences are established between the current camera observation and the next frame in the planned path (target) using XFeat \cite{potje2024xfeat} with LighterGlue. The matched 2D keypoints from the current observation are paired with 3D points reconstructed from the depth values at the target pose in the 3DGS. These depth values are back‑projected into 3D camera coordinates using the target pose intrinsics and then transformed into the global coordinate frame with the target pose extrinsics, yielding the 2D–3D correspondences required for PnP‑RANSAC pose estimation. These correspondences are then used by PnP \cite{hartley2003multiple} with RANSAC \cite{fischler1981random} to estimate the camera pose of the current observation within the 3DGS. The solution is then refined via Gauss‑Newton non‑linear minimization, yielding a final $4 \times 4$ transformation matrix that aligns the current observation to the next frame in the planned path. This matrix is used in a simple kinematic model for the Jackal robot that cannot move sideways: rotate to face the goal, translate forward, and rotate in place to match the target orientation.

\vspace{-0.55cm}
\paragraph{\bf GUI} A GUI was developed to make YOPO-Nav both intuitive and interpretable (see Fig.~\ref{fig:yopo_nav_gui}). The interface provides transparency into the YOPO-Nav's decisions and progress, while also supporting human intervention when needed. At any point, a human operator (using a controller connected locally or remotely) can issue discrete corrections of \SI{0.25}{m} forward/backward or $15^\circ$ left/right rotations, where the GUI helps to determine when such corrections are appropriate. Following interventions, YOPO-Nav resumes autonomous traversal by re-localizing the Jackal robot using YOPO-Loc and updating its position within the planned path.
\vspace{-0.1cm}
\section{Experiments and Results}
\label{sec:exper_results}

\subsection{Comparisons to SOTA}
\label{subsec:sota_comparisons}
\paragraph{\bf Baselines}
As a preliminary experiment, we tested replaying the controller actions from a YOPO-Campus trajectory on a physical robot in the same environment. As expected, this naive approach failed within the first few actions due to drift and error accumulation. To establish a benchmark, we compare YOPO-Nav to other image-goal navigation methods, namely ViNT \cite{shah2023vint} and NoMad \cite{sridhar2024nomad}, using real-world robot deployment. Both ViNT and NoMad boast generalizable navigation policies that are intended to work as zero-shot foundation models. While ViNT supports fine-tuning, we do not fine-tune YOPO-Loc on our own data, so we elect to compare ViNT and NoMad zero-shot. We run both algorithms using the author's implementation and their default navigation parameters. 

\vspace{-0.45cm}
\paragraph{\bf Experimental Setup} 
We evaluate YOPO-Nav on image-goal navigation across several different distance thresholds, with distances ranging from \SI{1.5}{m} to \SI{12}{m}. We evaluate three sample exploration videos in our dataset, and following CityWalker \cite{liu2024citywalker}, we evaluate the performance of the two algorithms on each trajectory at each distance threshold for 8-10 trials. These three exploration videos were chosen based on the trajectory shape and visual differences; they vary in the relative level of foliage, the number of buildings in the scene, and the number of turns. We deploy the robot in the same environment as the exploration videos, in which the scene appearance has significantly (i.e. tested in winter 2025, versus summer and fall 2024 when the dataset was collected). Minimal pedestrian traffic was also present during evaluation. Each algorithm was initialized with a modified version of the evaluation trajectory, in which the rotations at intersections (see Section \ref{subsec:yopo-campus}) were removed, and an image in the trajectory was specified as the goal. 

\vspace{-0.25cm}
\paragraph{\bf Metrics}
To measure performance we use success rate, where success is defined as the navigation algorithm reaching the location where the goal image was taken and recognizing it as the goal. Failure cases end the trial; failure cases include the navigation algorithm taking incorrect actions (i.e. turning right instead of left at an intersection), colliding with an object, or not recognizing the target location as the goal.

\vspace{-0.25cm}
\paragraph{\bf Real-World Deployment} 
Evaluation was conducted using a Clearpath Jackal UGV robot, whose onboard computer (i3‑9100TE CPU, no GPU) required a low‑latency link to a remote GPU workstation; we implemented this using s2n‑quic \cite{s2n-quic}, AWS’s Rust implementation of the QUIC protocol, selected for its broad compatibility, modern features (TLS 1.3, post‑quantum cryptography), and mutual TLS authentication over the public campus network. To optimize throughput, we built minimal client‑server scripts for sending and receiving strings and images. Interoperability between Rust and Python was achieved through PyO3 \cite{PyO3}. On the Jackal, the Python code queried ROS topics (RealSense D435 and velocity controller) while Rust code managed communication; on the remote workstation, Python code ran YOPO‑Nav while Rust code  again handled communication. Persistent connectivity across campus was maintained using a MiFi X Pro 5G hotspot on Calyx Institute’s unlimited mobile data plan.

\begin{table}[b!]
  \centering
  \scriptsize
  \setlength{\tabcolsep}{2pt}
  \scalebox{1.20}{
    \begin{tabular}{@{}lcccc@{}}
      \toprule
      Method & SR@\SI{1.5}{m} ↑ & SR@\SI{3}{m} ↑ & SR@\SI{7}{m} ↑ & SR@\SI{12}{m} ↑ \\
      \midrule
      ViNT \cite{shah2023vint} & 0.94 & 0.67 & 0.63 & 0.42 \\
      NoMaD \cite{sridhar2024nomad} & 0.83 & 0.53 & 0.53 & 0.20 \\
      \midrule
      YOPO-Nav & \textbf{1.0} & \textbf{0.94} & \textbf{0.81} & \textbf{0.75} \\
      \bottomrule
    \end{tabular}
  }
  \caption{Comparison of image-goal success rate at different trajectory lengths across state-of-the-art navigation methods, with best performing results in bold. YOPO-Nav significantly outperforms zero-shot transformer-based models at both short and long trajectories in real-world campus environments.}
  \label{tab:yopo-metrics}
\end{table}

\begin{table*}[t!]
  \centering
  \scriptsize
  \setlength{\tabcolsep}{2.5pt}
  \scalebox{1.15}{
    \begin{tabular}{@{}l|ccccccc@{}}
      \toprule
      Path Type & Length (\SI{}{m}) & Human Interventions & Actions per Int. & Robot Actions & Time (min) & GT Actions & GT Time (min) \\
      \midrule   
      Straight Trajectory & 222.6 & 5 & 17.6 & 2280 & 25.8 & 908 & 15.8 \\
      Varying Ground Type & 113.6 & 7 & 18.7 & 1281 & 10.4 & 535 & 3.3 \\
      Construction Area & 167.0 & 15 & 16.5 & 1968 & 21.1 & 787 & 4.5 \\
      Seasonal Changes & 92.4 & 8 & 18 & 1245 & 13.4 & 446 & 4.9 \\
      Combined Trajectory/Downward Slope & 87.2 & 3 & 14 & 1302 & 14.2 & 572 & 4.6 \\
      \bottomrule
    \end{tabular}
  }
  \caption{
  Summary of collected metrics testing image-goal navigation using the YOPO-Nav algorithm with human interventions and accumulated human actions at each intervention. Metrics include length of the exploration path, the number of human interventions for a successful navigation, the mean number of controller micro-actions per intervention (Actions per Int.), the number of total robot actions to reach the goal image, time to completion, the ground truth number of actions (during the teleoperation stage), and the ground truth time. Notice the simple trajectories (e.g. Straight Trajectory) require fewer human interventions, while exploration paths with large scene changes (e.g. Construction Area) require more human interventions.}
  \label{tab:yopo-interventions}
\end{table*}

\vspace{-0.25cm}
\paragraph{\bf Implementation Details} 
The visual place recognition model used a DINOv3 (ViT-L) \cite{simeoni2025dinov3} backbone with the final two layers unfrozen for fine-tuning. A BoQ \cite{ali2024boq} aggregator consisting of two layers (512 features, 128 learnable queries) produced 4096 descriptors from images that were resized to 224×224 during training and validation. The model was trained on GND \cite{liang2025gnd} for 30 epochs with a batch size of 64, learning rate of $1 \times 10^{-4}$ (with a warmup factor of 0.1), and weight decay of $1 \times 10^{-4}$. Frames from each exploration video were processed in AnySplat \cite{jiang2025anysplat} sequentially in batches of 55, to balance accuracy and quality within VRAM limits of the NVIDIA RTX 4090, at $448 \times 336$ resolution and saved as .ply files without spherical harmonics; SOG compression reduced file sizes from 150–250 MB to 25–40 MB in 15–30 seconds. Reconversion to .ply was performed during inference. Pose estimation was performed using OpenCV’s \cite{opencv_library} PnP \cite{hartley2003multiple} with RANSAC \cite{fischler1981random}, configured to use SQPNP with 500 iterations, a 5‑pixel reprojection threshold, and a 0.99 confidence threshold. Ground plane estimation was restricted to the lower $30\%$ of the renders at each known pose in the 3DGS, subsampled to 20,000 points, and filtered to the closest fiftieth percentile of depths, with RANSAC using 500 iterations and an inlier threshold of 0.01 to fit the plane.

\vspace{-0.30cm}
\paragraph{\bf Results}
Table \ref{tab:yopo-metrics} shows the relative success rates of YOPO-Nav compared to ViNT \cite{shah2023vint} and NoMad \cite{sridhar2024nomad} at each distance threshold. Notice that YOPO-Nav outperforms its competitors at short trajectories, and significantly outperforms its competitors at long trajectories. It is unlikely that a lack of appropriate training data is the cause of this performance difference, since ViNT and NoMad include similar campus environments in their training data such as SCAND \cite{karnan2022socially} and Berkeley \cite{shah2022viking}. 

Differences in scene dynamics (e.g., seasonal variation, lighting conditions, and activity such as pedestrian traffic) can affect the performance of end‑to‑end solutions such as ViNT and NoMad. Our method, however, is robust to these variations and other dynamics that differ from the exploration video, so long as they do not dominate the scene, and it produces stronger success rate metrics under similar navigation conditions.

\subsection{Human Interventions}
\label{subsec:human_intervention_exprs}

\paragraph{\bf Motivation}
Operating robots in real-world environments is challenging. Intuitively, human-in-the-loop intervention can help robots achieve their navigation goals. When such intervention can occur without retraining the robot, interaction becomes more seamless, enabling navigation to function as a true human–robot collaboration. Human assistance has generally proven effective in many recent works \cite{saunders2018trial, chen2020ask, lee2021pebble, christiano2017deep}. Building on this idea, we enable a companion human to provide simple interventions via a controller when the robot deviates from its intended path. Using this setup, we evaluated YOPO‑Nav on five full‑length paths in YOPO-Campus to examine how well the algorithm can scale while keeping human interventions to a minimum.

\vspace{-0.3cm}
\paragraph{\bf Experimental Setup}
We used the same real-world deployment and implementation details from earlier. For these trials, we initialize YOPO-Nav with an unmodified exploration video from YOPO-Campus, and place the Clearpath Jackal robot in a similar starting point as in the exploration video. Each trajectory was tested for one trial. A human-in-the-loop monitored each trial and teleoperated the robot with small discrete actions, effectively a nudge, if it went off the sidewalk, was about to collide, or could not determine the next action. Interventions to prevent potential collisions were overly cautious. We choose to test YOPO-Nav using five trajectories that contain varying scenes, trajectories, and challenges:
\noindent
(1) \textit{Straight Trajectory}: a trajectory between two buildings where the primary action to reach the image goal is to move forward in a straight line.
\noindent
(2) \textit{Varying Ground Type}: a trajectory with varying ground types including brick and cement, as well as multiple turns.
 \noindent
(3) \textit{Construction Area}: a trajectory whose scene changed significantly due to construction in the area.
\noindent
(4) \textit{Seasonal Changes}: a trajectory with seasonal foliage in the scene which changed between the exploration video and the trial.
\noindent
(5) \textit{Combined Trajectory/Downward Slope}: two trajectories stitched together into a longer video, with a sloping ground plane present in a section.

\vspace{-0.3cm}
\paragraph{\bf Metrics}
We collect four distinct metrics to evaluate the effect of human interventions on the performance of YOPO-Nav:
\noindent
(1) \textit{Human Interventions}: the number of manual actions by a human necessary to successfully navigate to the goal image. A human intervention is defined as any instance in which the human‑in‑the‑loop collaborator took control of the mobile robot during one of the three failure cases previously discussed.
\noindent
(2) \textit{Actions per Intervention}: the mean number of controller action inputs by the human collaborator per each intervention. Following the same discrete actions as the YOPO-Campus dataset, each controller action is limited to \SI{0.25}{m} forward or backward, or $15^\circ$ left or right.
\noindent
(3) \textit{Robot Actions}: the number of robot actions taken to successfully navigate to the goal image. Human intervention actions were not included in the robot actions metric.
\noindent
(4) \textit{Time}: the amount of time the robot took to successfully navigate to the goal image in minutes.

These metrics are listed along with ground truth data of the exploration path, including:
\noindent
(1) \textit{Length}: the length of the exploration path in meters.
\noindent
(2) \textit{Ground Truth (GT) Actions}: the number of controller input actions taken by the human teleoperator in the exploration path.
\noindent
(3) \textit{GT Time (min)}: the amount of time it took for the human teleoperator to complete the exploration path.

\vspace{-0.35cm}
\paragraph{\bf Results}
Table \ref{tab:yopo-interventions} shows the human intervention evaluation metrics for each path type trial. The length of the path generally does not imply a higher number of human interventions. This implies that YOPO-Nav is resilient to potential noise present in a trial that would be compounded over long trajectories. The number of human interventions correlates with the difficulty of the task, and is higher in cases of scene changes between the exploration video and the trial environment. YOPO‑Nav is limited in its ability to adapt to changes that overly dominate the scene, as shown in the Construction Area case, since its scene representation is a snapshot at a fixed point in time; as shown in the other cases, however, YOPO-Nav remains largely effective when evaluated on paths whose challenges do not involve relatively extreme scene change. Despite this shortcoming, YOPO-Nav can successfully complete image-goal navigation tasks over long trajectories with a human-in-the loop collaborator.
\vspace{-0.15cm}
\section{Conclusion}
We present a novel scene representation for visual navigation, composed of a graph of 3DGS that partition an environment into small, manageable chunks. We introduce the YOPO-Nav framework using a 3DGS graph and a VPR model to perform image-goal navigation. Our results show that YOPO-Nav outperforms zero-shot foundation models and performs successful visual navigation on long trajectories with limited human-in-the-loop intervention.

The YOPO-Campus dataset has a {\bf companion dataset} {\it Ego-Campus} \cite{john2025egocampusegocentricpedestrianeye}, with egocentric video and eye-gaze from pedestrians walking identical campus paths (shown in Fig.~\ref{fig:yopo_campus_bev} wearing Project Aria glasses \cite{engel2023project}. While YOPO-campus shows the robot view, Ego-campus shows a pedestrian view and captures eye-gaze. Together, these datasets provide an important data resource for future studies of navigation in human-robot systems. 

\vspace{-0.15cm}
\section*{Acknowledgments}
 This work was supported by the NSF-NRT grant: Socially Cognizant Robotics for a Technology Enhanced Society (SOCRATES), No. 2021628, and by the National Science Foundation (NSF) under Grant Nos. CNS-1901355, CNS-1901133.
{
    \small
    \bibliographystyle{ieeenat_fullname}
    \bibliography{main}
}

\end{document}